# Decision-Making under Ordinal Preferences and Comparative Uncertainty[†]

Didier Dubois, Hélène Fargier, Henri Prade
Institut de Recherche en Informatique de Toulouse – Université Paul Sabatier
118 route de Narbonne – 31062 Toulouse Cedex 4 – France
Email: {dubois, fargier, prade}@irit.fr

## Abstract

This paper proposes a method that finds a preference relation on a set of acts from the knowledge of an ordering on events describing the decision-maker's uncertainty and an ordering of consequences of acts, describing the decision maker's preferences. However, contrary to classical approaches to decision theory, this method does not resort to any numerical representation of utility nor uncertainty and is purely ordinal. It is shown that although many axioms of Savage theory can be preserved and despite the intuitive appeal of the ordinal method, the approach is inconsistent with a probabilistic representation of uncertainty. It leads to the kind of uncertainty theory encountered in non-monotonic reasoning (especially preferential and rational inference). Moreover the method turns out to be either very little decisive or to lead to very risky decisions, although its basic principles look sound. This paper raises the question of the very possibility of purely symbolic approaches to Savage-like decision-making under uncertainty and obtains preliminary negative results.

## 1 INTRODUCTION

In Savage decision theory (Savage, 1972), a decision problem is cast in the framework of a set of states (of the world) and a set of consequences. An act is viewed as a mapping f from the finite state space S to the consequence set X, namely in each state s the result of f is $f(s) \in X$. Savage is interested in the following problem: Starting from a user-driven preference relation over acts, i.e., $(X^S, \geq)$, and axioms that such a preference relation should obey if the decision-maker is "rational", construct a relation representing uncertainty on the subsets of S, and a preference relation on the consequences. If $(X^S, \geq)$ satisfies suitable properties then the uncertainty on S can be represented by a probability distribution p, the preference relation on X by a utility function u, and the preference relation on acts by expected utility:

$$f \geq g \Leftrightarrow \sum_{s \in S} p(s)u(f(s)) \geq \sum_{s \in S} p(s)u(g(s)).$$

In this research, we are first interested in the converse problem, namely: if an uncertainty relation on events in $2^S$ is given, and a preference relation on consequences is given as well, how can a preference relation on acts be recovered with a purely symbolic approach? This problem is realistic because a decision maker is not necessarily capable of describing his state of uncertainty by means of a probability distribution, nor may he be able to quantify his preferences on X. A natural technique that solves this problem is described. Applying this technique to a comparative probability relation on events we get a non-transitive relation on acts. However this anomaly is avoided if we assume a possibility ordering on events.

A characterization of this kind of approach is obtained by preserving the basic axioms of Savage augmented with an axiom sanctioning the proposed ordinal act-ordering technique, but without assuming transitivity of indifference between acts. However as shown here, the uncertainty relation obtained in this general setting is closely related to preferential inference in nonmonotonic reasoning and the decision guidelines obtained in this setting are questionable.

## 2 FROM UNCERTAINTY AND PREFERENCE RELATIONS TO AN ORDERING OF ACTS

This section introduces a natural technique to compute a preference relation on acts from a purely symbolic perspective. Let $(2^S, \geq_U)$ be an uncertainty relation on events. We assume that $\geq_U$ is a complete partial ordering, non-trivial ($S >_U \emptyset$), and faithful to deductive inference:

$$A \subseteq B \Rightarrow A \leq_U B.$$

As usual, let us define the indifference $\sim_U$ and the strict relation $<_U$ induced from $\geq_U$ by:

$A \sim_U B$ iff $A \leq_U B$ and $B \leq_U A$
$A <_U B$ iff $A \leq_U B$ and not $(B \leq_U A)$.

---

[†]This paper is a fully revised and extended version of a preliminary workshop paper (Dubois et al., 1997).

The preference on X is just a complete partial ordering $\leq_P$. It is assumed that X has at least two elements x and y s.t. $x >_P y$.

If no commensurability assumption can be made, a natural way of lifting ($\leq_U$, $\leq_P$) on $X^S$ is to assume that an act f is at least as promising as an act g iff the event $[f \geq g]$ formed by the disjunction of states in which f gives results at least as good as g, is at least as likely as the event $[g \geq f]$, formed by the disjunction of states in which g gives results at least as good as f. A state s is as a more promising state for f than for g iff $f(s) \geq_P g(s)$. Hence, we construct the preference between acts and the corresponding indifference and strict preference relations as:

**Definition (L).**
$f \geq g$ iff $[f \geq g] \geq_U [g \geq f]$ where $[f \geq g] = \{s, f(s) \geq_P g(s)\}$;
$f \sim g$ iff $f \geq g$ and $g \leq f$;
$f > g$ iff $f \geq g$ and not($g \leq f$).

This approach seems to be very natural, and is the first one that comes to mind when information is only available under the form of an uncertainty relation on events and a preference relation on consequences, *if the preference and uncertainty scales are not commensurate*.

The properties of the relations $\geq$, $\sim$, and $>$ on $X^S$ will depend on the properties of $\geq_U$. The most obvious choice for $\geq_U$ is a qualitative probability (e.g., Fishburn, 1986):

**Definition.** $\geq_U$ is a qualitative probability iff
i) $\geq_U$ is complete and transitive
ii) $S >_U \emptyset$, $\forall A \geq_U \emptyset$
iii) $A \geq \emptyset$, $A \cap (B \cup C) = \emptyset \Rightarrow (B \geq_U C \Leftrightarrow A \cup B \geq_U A \cup C)$.

A first negative result is that if $\geq_U$ is a qualitative probability ordering then relation $>$ in $X^S$ is not necessarily transitive. So the ordinal approach embodied in definition (L) cannot agree with Savage decision theory.

**Example.**
| States | s1 | s2 | s3 | s4 | s5 | s6 |
|---|---|---|---|---|---|---|
| Prob. | 2/9 | 1/9 | 1/9 | 2/9 | 1/9 | 2/9 |
| f | 5 | 100 | 0 | 0 | -10 | -10 |
| g | 0 | -15 | 100 | -10 | 0 | 10 |
| h | -25 | 0 | -40 | 20 | 40 | 0 |

It is easy to verify that :

P([f≥g])=P({s1,s2,s4})=5/9 >P([g≥f])=P({s3,s5,s6})=4/9
P([g≥h])=P({s1,s3,s6})=5/9 > P([h≥g])=P({s2,s4,s5})=4/9
P([f≥h])=P({s1,s2,s3})=4/9 >P([h≥f])=P({s4,s5,s6})=5/9.

Hence f>g>h but h>f. Note that the intransitivity result does not depend on the figures in the table insofar as the sign of the utility values is respected. This situation can be viewed as an analog of the Condorcet paradox in social choice, here in the setting of decision under uncertainty.

Thinking about an alternative approach, the simplest type of qualitative uncertainty ordering are those induced by necessity measures (Dubois, 1986):

**Definition.** $\geq_N$ is a qualitative necessity ordering iff
i) $\geq_N$ is complete and transitive
ii) $S >_N \emptyset$, $\forall A \geq_N \emptyset$
iii) $B \geq_N C \Rightarrow A \cap B \geq_N A \cap C$.

They are also epistemic entrenchments in the sense of Gärdenfors (Dubois and Prade, 1991), and comparative possibility orderings of Lewis (1973). From now on, we assume that $\geq_U = \geq_N$. Any necessity ordering can be represented by a necessity function on an ordinal scale:

**Property** (Dubois, 1986). $\exists N$ such that: $B \geq_N C \Leftrightarrow N(B) \geq N(C)$ where $N(F) \in [0,1]$ and $N(F \cap G) = \min(N(F), N(G))$ whatever F and G.

Necessity orderings are simple since that they can be derived from a complete partial ordering on states only, and can be encoded as a possibility distribution. We can now write in accordance with definition (L) :

$$f \geq g \Leftrightarrow N([f \geq g]) \geq N([g \geq f]). \qquad (1)$$

In terms of possibility measures (Zadeh, 1978), $\Pi(A) = 1 - N(\bar{A})$, then:
**Property.** $f \geq g \Leftrightarrow \Pi([f > g]) \geq \Pi([g > f])$ where $[f > g] = \{s, f(s) >_P g(s)\}$.

## 3  PROPERTIES OF THE POSSIBILITY-DRIVEN RELATION ON ACTS

In accordance with Savage, we use the following definitions:

**Definitions.**
- $f_{|A} \cup g_{|\bar{A}}$ is the act s.t. $(f_{|A} \cup g_{|\bar{A}})(s) = f(s)$ if $s \in A$,
  $= g(s)$ otherwise.
- $f \geq_A g$ iff $\forall h\ f_{|A} \cup h_{|\bar{A}} \geq g_{|A} \cup h_{|\bar{A}}$
- A null iff $\forall f, g$, $f \geq_A g$
- $\forall x, y \in X$, $x \geq y$ iff $fx \geq fy$ where $fx: \forall s \in S\ fx(s)=x$ and $fy: \forall s \in S\ fy(s)=y$

$f_{|A} \cup g_{|\bar{A}}$ is like a lottery whose result is state dependent. $f \geq_A g$ denotes the notion of conditional preference over acts. A null event is a set of states in which the result of acts does not matter. The reason for this indifference is that the agent considers that this act will never happen. The last point identifies constant acts fx to their (unique, state-free) consequence, so that $X \subseteq X^S$. In case of necessity measures on S, the following properties of $(X^S, \geq)$ can be established, where $\geq$ is defined by (1):

**Property.**    $\forall x, y \in X$    $x \geq y \Leftrightarrow x \geq_P y$.

**Property.**  $f \geq_A g \Leftrightarrow N([f \geq g] \cup \bar{A}) \geq N([g \geq f] \cup \bar{A})$
$\Leftrightarrow \Pi([f > g] \cap A) \geq \Pi([g > f] \cap A)$.

**Proof (sketch).** Let us denote $B = [f_{|A} \cup h_{|\bar{A}} \geq g_{|A} \cup h_{|\bar{A}})]$ and $C = [f_{|A} \cup h_{|\bar{A}} \leq g_{|A} \cup h_{|\bar{A}}]$.

Remark that, by definition, whatever h: $B = [f \geq g] \cup \bar{A}$ and $C = [g \geq f] \cup \bar{A}$. □

Hence: $N(B) \geq N(C) \Leftrightarrow f \geq_A g$.

**Property.**  A null $\Leftrightarrow \Pi(A) = \Pi(\emptyset)$.

**Property P1' (Restricted Savage's P1)**
- $(X^S, \geq)$ is complete : $f \geq g$ or $g \geq f$ for all pairs of acts;
- The indifference relation ~ is reflexive, symmetrical;
- The strict relation > on $X^S$ is a transitive, irreflexive, partial order;
- $f > g$ and $g \sim h \Rightarrow f \geq h$.

**Proofs.** the only difficulty is to prove that > is transitive. Let f,g,h be three acts s.t. $f > g$, $g > h$, $h \geq f$, i.e.
$$\Pi([f < g]) < \Pi([f > g])$$
$$\Pi([g < h]) < \Pi([g > h])$$
$$\Pi([f < h]) \geq \Pi([f > h]). \quad (i)$$

Let us partition S in 13 subsets:

$A = \{s / f(s) = g(s) = h(s)\}$  $B = \{s / f(s) = g(s) < h(s)\}$
$C = \{s / f(s) = g(s) > h(s)\}$  $D = \{s / f(s) = h(s) < g(s)\}$
$E = \{s / f(s) = h(s) > g(s)\}$  $F = \{s / g(s) = h(s) < f(s)\}$
$G = \{s / g(s) = h(s) > f(s)\}$  $H = \{s / f(s) < g(s) < h(s)\}$
$I = \{s / f(s) < h(s) < g(s)\}$  $J = \{s / g(s) < f(s) < h(s)\}$
$K = \{s / h(s) < f(s) < g(s)\}$  $L = \{s\ g(s) < h(s) < f(s)\}$
$M = \{s / h(s) < g(s) < f(s)\}$.

The system of inequations (i) can be rewritten as:

$\max(e,f,j,l,m) > \max(d,g,h,i,k)$ (ii-1)
$\max(c,d,i,k,m) > \max(b,e,h,j,l)$ (ii-2)
$\max(b,g,h,i,j) \geq \max(c,f,k,l,m)$ (ii-3)

(where a is the possibility degree of A, etc.). It is inconsistent. Indeed, from ii-1 and ii-2, one must have $\max(f,c,m)$ greater than all of b,d,e,g,h,i,j,k,l, and this is in contradiction with ii-3. □

Postulate P1 of Savage states that the weak preference $\geq$ on acts should be a complete and transitive relation. So the only difference between P'1 and Savage's P1 is the lack of transitivity of the indifference relation. This lack of transitivity can be observed in very simple cases, where only 3 states and 2 consequences are defined. Our formalism also differs from Lehmann's (1996) who assumes that $\geq$ is not complete but the indifference is transitive. Non-transitive indifference is natural if $f \sim f'$ means closeness. But our formalism satisfies some of his axioms, in particular :

**Property.** If $f \geq_A g$ and $f \sim_{\bar{A}} g \Rightarrow f \geq g$. (U)

This easily checked property, which sounds like an unanimity principle, is usually deduced from Savage's axiomatic decision theory (however, using the transitivity of $\geq$).

**Property P2.** The preference relation $\geq$ obeys the sure thing principle:
$\forall f, g, h, h'\ f_{|A} \cup h_{|\bar{A}} \geq g_{|A} \cup h_{|\bar{A}}$
$$\Leftrightarrow f_{|A} \cup h'_{|\bar{A}} \geq g_{|A} \cup h'_{|\bar{A}}$$

**Sketch of proof.** This is because $[f_{|A} \cup h_{|\bar{A}} \geq g_{|A} \cup h_{|\bar{A}}] = \bar{A} \cup ([f \geq g] \cap A)$ does not depend on h.

The sure thing principle is the cornerstone of Savage decision theory, and it is instrumental in getting a probabilistic representation of uncertainty. The fact that it is compatible here with a possibilistic representation is due to the fact that indifference between acts is not transitive.

**Property P3 (Compatibility of conditional preference with constant acts).**
$\forall A \subseteq S$, A not null, $fx \geq_A fy \Leftrightarrow x \geq y$.

This is Savage Postulate P3. It has intuitive appeal in any framework where the sure thing principle applies.

**Proof.**
- $x \sim y \Leftrightarrow x \sim_P y \Rightarrow [fx \geq fy] = [fy \geq fx] = S$ thus:
$\forall A, N([fx \geq fy] \cup \bar{A}) = N([fy \geq fx] \cup \bar{A}) \Leftrightarrow$
$$\forall A\ fx \sim_A fy \quad (i)$$
- $x > y \Leftrightarrow x >_P y \Rightarrow [fx > fy] = S$ and $[fy \geq fx] = \emptyset$ thus:
$\forall A, [fx > fy] \cap A = A$ and $[fy > fx] \cap A = \emptyset$
A is not null: $\Pi(A) > \Pi(\emptyset)$ and $\Pi([fx > fy] \cap A) = \Pi(A) > \Pi([fy > fx] \cap A)$. Hence: $fx >_A fy$ (ii)
- By (i) and (ii): $x \geq y \Rightarrow fx \geq_A fy$. Moreover, by (ii) contraposed: $fx \leq_A fy \Rightarrow x \leq y$. Exchanging x and y: $fy \leq_A fx \Rightarrow y \leq x$. □

$(S, \geq_U)$ and $(X, \geq_P)$ induce $(X^S, \geq)$; let us now reproject $X^S$ on $2^S$ by considering 2-consequence acts $\omega(C)^{x,y}$ where $x > y$ and $\omega(C)^{x,y}(s) = x$ if $s \in C$, $\omega(C)^{x,y} = y$ otherwise

**Property P4.** Restriction to 2-consequence acts.
$\forall x, y, x', y'$ s.t. $x > y$, $x' > y'$, $\omega(A)^{x,y} \geq \omega(B)^{x,y}$
$$\Leftrightarrow \omega(A)^{x',y'} \geq \omega(B)^{x',y'}.$$

**Proof.** Since $x > y$ and $x' > y'$:
$[\omega(C)^{x,y} \geq \omega(B)^{x,y}] = C \cup \bar{D}$ does not depend on $(x,y)$
Thus: $\omega(A)^{x,y} \geq \omega(B)^{x,y} \Leftrightarrow N(A \cup \bar{B}) \geq N(\bar{A} \cup B)$
$$\Leftrightarrow \omega(A)^{x',y'} \geq \omega(B)^{x',y'} \quad \square$$

This property which is exactly Savage Postulate P4 enables events A and B to be consistently compared by fixing $x > y$ arbitrarily when selecting $\omega(A)^{x,y}$ and $\omega(B)^{x,y}$. It operates a restriction from $X^S$ to $2^S$.

**Definition.**  $A \geq B \Leftrightarrow \forall\ x > y, \omega(A)^{x,y} \geq \omega(B)^{x,y}$

Lehmann (1996) defines $A \geq B$ in the same way, but drops P4.

**Properties.**

$$A \geq B \Leftrightarrow \exists\, x > y,\ \omega(A)^{x,y} \geq \omega(B)^{x,y} \text{ (due to P4)}$$
$$\Leftrightarrow N(\bar{B} \cup A) \geq N(\bar{A} \cup B)$$
$$\Leftrightarrow \Pi(A \cap \bar{B}) \geq \Pi(\bar{A} \cap B).$$

So, although different from $\geq_N$, the relation $\geq$ obtained via lifting followed by restriction can be expressed in terms of the necessity ordering $\geq_N$, and it refines it.

Finally, postulate P5 of Savage (X has at least two elements x and y s.t. x>y) will be trivially satisfied in any non-trivial decision theory.

## 4  THE POSSIBILISTIC LIKELIHOOD RELATION

The relation $\geq$ on $2^S$ is closely related to the necessity and possibility orderings induced by N (and its dual $\Pi$); more specifically it is a refinement of both orderings:

**Properties.**   • $N(A) > N(B) \Rightarrow A > B$
•  $\Pi(A) > \Pi(B) \Rightarrow A > B$.

But it may happen that $N(A) = N(B)$ and $A>B$ or that $\Pi(A) = \Pi(B)$ and $A>B$. Notice that $\sim$ on $2^S$ is not transitive since:

$$\Pi(A \cap \bar{B}) = \Pi(\bar{A} \cap B)$$
$$\Pi(B \cap \bar{C}) = \Pi(\bar{B} \cap C) \not\Rightarrow \Pi(A \cap \bar{C}) = \Pi(\bar{A} \cap C).$$

In fact, the indifference $\sim$ on $2^S$ is transitive when considering only $\Pi$-mutually exclusive events: $\Pi(A \cap B) = \min(\Pi(A), \Pi(B))$ means that A and B are not $\Pi$-mutually exclusive according to $\Pi$, and conversely A and B are said to be $\Pi$-mutually exclusive if $\Pi(A \cap B) \neq \min(\Pi(A), \Pi(B))$ that is if $\Pi(A \cap B) < \min(\Pi(A \cap \bar{B}), \Pi(\bar{A} \cap B))$.

**Property.**  $\Pi(A) \geq \Pi(B) \Leftrightarrow A \geq B$ iff A and B are $\Pi$-mutually exclusive.

In other terms, $\geq$ refines $\geq_N$ and the dual possibilistic ordering $\geq_\Pi$ for not $\Pi$-mutually exclusive events.

The partial ordering $>$ on events is a special case of the so-called "discrimax" relation between vectors (Dubois et al., 1996; Fargier et al., 1993):

$$A > B \Leftrightarrow \max_{i \in \mathcal{D}(A,B)} a_i \geq \max_{i \in \mathcal{D}(A,B)} b_i$$

where $a_i = \Pi(\{s_i\})$ if $s_i \in A$ and $\mathcal{D}(A,B) = \{i, a_i \neq b_i\}$.

Finally, since $\geq$ on $X^S$ satisfies Savage's sure thing principle, we have the additivity condition (iii) of qualitative probabilities and the autoduality property:

$$A \geq B \Leftrightarrow \bar{A} \leq \bar{B}.$$

However, since the indifference relation between events is not transitive, this relation cannot be represented by a probability. Because the relation on events is closely related to necessity and possibility orderings, we shall call it possibilistic likehood.

## 5  LINK WITH NONMONOTONIC REASONING

To summarize, the possibilistic likelihood relation obtained in Section 3 is a qualitative probability ordering, but for the transitivity of indifference. The derived possibilistic likehood relation $\geq$ also verifies properties that are NOT satisfied by qualitative probabilities:

**Properties.**
• $A \cap (B \cup C) = \emptyset$ and $A \geq B$ and $A \geq C \Rightarrow A \geq B \cup C$
• $A \cap (B \cup C) = \emptyset$ and $A > B$ and $A > C \Rightarrow A > B \cup C$
• if A, B, C are pairwise disjoint then $A \cup B > C$ and $A \cup C > B$ imply $A > B \cup C$.

The latter property is closely related to one of the characteristic properties for uncertainty functions that represent acceptance (Dubois and Prade, 1995b; Friedman and Halpern, 1996), i.e., a function inducing orderings for which for any not empty set A, the set $\{B, A \cap B > A \cap \bar{B}\}$ (the set of propositions accepted by $>$ when A is true) is deductively closed. A nonmonotonic consequence relation can be derived from such acceptance orderings as :

$$A \mathrel{|\!\sim} B \Leftrightarrow A \cap B > A \cap \bar{B}.$$

The comparative possibility ordering $\Pi(A) > \Pi(B)$ can also be related to nonmonotonic inference $A \mathrel{|\!\sim} B$ which expresses that $\Pi(A \cap B) > \Pi(A \cap \bar{B})$ for a possibility measure $\Pi$. Possibility theory is closely related to preferential inference in nonmonotonic reasoning, as defined by Kraus, Lehmann and Magidor (1990). The properties of preferential inference are

RR : $A > \emptyset \Rightarrow A \mathrel{|\!\sim} A$.
AND: $A \cap C > A \cap \bar{C}$ and $A \cap B > A \cap \bar{B} \Rightarrow A \cap B \cap C > A \cap (\bar{C} \cup \bar{B})$
OR: $A \cap C > A \cap \bar{C}$, $B \cap C > B \cap \bar{C} \Rightarrow (A \cup B) \cap C > (A \cup B) \cap \bar{C}$
RW: $B \subseteq C$, $A \cap B > A \cap \bar{B} \Rightarrow A \cap C > A \cap \bar{C}$
CM: $A \cap B > A \cap \bar{B}$ and $A \cap C > A \cap \bar{C} \Rightarrow A \cap B \cap C > A \cap B \cap \bar{C}$.
CUT: $A \cap B > A \cap \bar{B}$ and $A \cap B \cap C > A \cap B \cap \bar{C} \Rightarrow A \cap C > A \cap \bar{C}$.

Then, an inference relation $\mathrel{|\!\sim}$ is preferential if and only if there exists a set of positive possibility measures[1] $\mathcal{F}$ such that (Dubois, Prade 1995a):

---
[1] A possibility measure is positive iff, $\forall A$, $\Pi(A) > \Pi(\emptyset)$.

$A \succ B$ iff $\forall \Pi \in \mathcal{F}, \Pi(A \cap \bar{B}) > \Pi(A \cap \bar{B})$.

In (Benferhat et al.,1992) it is shown that when contains a single possibility measure, the obtained consequence relation is characteristic of rational inference of Lehmann (Lehmann and Magidor, 1992), that is an inference which satisfies rational monotony on top of the above axioms:

**Axiom (RM):** $A \cap B > A \cap \bar{B}$ and $A \cap \bar{C} \le A \cap C \Rightarrow A \cap B \cap C > A \cap \bar{B} \cap C$.

Since the sets $A \cap B$ and $A \cap \bar{B}$ are disjoint it holds:

$$A \succ B \Leftrightarrow A \cap B > A \cap \bar{B} \Leftrightarrow \Pi(A \cap B) > \Pi(A \cap \bar{B})$$

where $>$ is the strict possibilistic likehood relation associated with $\Pi$. Conversely, it is easy to check that $A>B$ can be written as:

$$A \Delta B \succ \bar{B} \cup A$$

where $\Delta$ denotes the symmetric difference.

## 6 FROM AXIOMS ON PREFERENCE OVER ACTS TO A THEORY OF UNCERTAINTY

Since the five first of Savage axioms have been retrieved (up to the transitivity of indifference) in the previous investigation, it is interesting to start the other way round, namely given a preference relation $\ge$ on acts, that satisfies the above Savage-like properties, determine the relation on events induced by $\ge$ using Savage's P4 and the relation on consequences using Savage's P3 that enable the original relation $\ge$ on acts to be recovered using the lifting procedure (L). Doing so, the approach is much more general: the assumption that uncertainty is represented by possibility measures is no longer necessary since as in the tradition of decision theory, the uncertainty representation now comes from the properties of the preference over acts only.

Let $X^S$ be a set of acts equipped with a preference relation $\ge$ such that

- P'1: $(X^S, \ge)$ is complete: $d_1 \ge d_2$ or $d_2 \ge d_1$ for all pairs of acts,
  $(X^S, >)$ is a transitive, irreflexive, partially ordered set,
  $(X^S, \sim)$ defines a symmetrical and reflexive relation.
- P2: $(X^S, \ge)$ satisfies the sure thing principle.
- P3: $(X^S, \ge)$ satisfies Savage's axiom of compatibility with constant acts.
- P4: $(X^S, \ge)$ satisfies Savage's P4 axiom about 2-consequence acts.
- P5': $\exists x, y, z$ three constant acts such that $x>y>z$.

NB: Note that Savage's P5 has been strengthened to 3 consequence sets and P'1 reflects the lack of transitivity of indifference. Other postulates of Savage are more technical and not as convincing. One pertains to infinite state spaces. The other is meant to cope with infinite consequence sets.

From the above set of axioms, we get the following properties of the relation on events (most of these proofs are easy and omitted to the sake of brevity):

**Properties.**
- $S > \emptyset$ (Pr. 6.1)
- $\forall A \subseteq S, A \ge \emptyset$ (Pr. 6.2)
- $A \cap (B \cup C) = \emptyset \Rightarrow (B \ge C \Leftrightarrow A \cup B \ge A \cup C)$
  (Pr. 6.3 additivity)
- $A > B \Leftrightarrow A \cap \bar{B} > \bar{A} \cap B$  • $A \ge B \Leftrightarrow A \cap \bar{B} \ge \bar{A} \cap B$
  Pr. 6.4)
- $A > B \Leftrightarrow \bar{A} < \bar{B}$  • $A \ge B \Leftrightarrow \bar{A} \le \bar{B}$  Pr 6.5: auto-duality)
- $A \subseteq B \Rightarrow B \ge A$   (Pr 6.6: coherence with set inclusion)
- if $A \subseteq B$ then $B - A > \emptyset \Leftrightarrow B > A$ (Pr. 6.7)
- $A$ null $\Leftrightarrow A \sim \emptyset$ (Pr. 6.8)
- $A > C, A \subseteq B, B - A > \emptyset \Rightarrow B > C$ (Pr. 6.9)
- $A \sim C, A \subseteq B, B - A > \emptyset \Rightarrow B \ge C$ (Pr. 6.10)
- $A > C, A \subseteq B \Rightarrow B \ge C$ (Pr. 6.11).

The properties (6.1- 6.6) indicate the close links between the uncertainty theory generated and qualitative probability; (6.7) and (6.8) show the importance of non-null events. (6.9), (6.10), (6.11) show that the strict ordering of events does not propagate totally via inclusion of sets, nor via indifferent events.

### 6.1 CONSEQUENCES OF THE LIFTING AXIOM

Let us now add the lifting Axiom to our set of axioms characterising the preference on acts :

$$L: f \ge g \Leftrightarrow [f \ge g] \ge [g \ge f].$$

Clearly this assumption is very strong since it prescribes a particular behavior for the decision-maker. Under these conditions we can show that:

**Properties.**
- $\forall s \in S, \{s\} > \emptyset \Rightarrow \forall A$ s.t. $s \in A, A > \emptyset$ (Pr. 6.12)
- $A \sim \emptyset \Rightarrow \forall s \in A \{s\} \sim \emptyset$ (Pr. 6.13)
- If at least two different states of S are not null, the preference on X is a complete preorder (but the ordering of states in S may be partial) (6.14).

**Proof of 6.14.** By (P1'), we know that $\ge$ is complete and $>$ transitive. We now have to show that the indifference relation $\sim$ that one can define from $\ge$ is also transitive on constant acts.

Suppose that x,y, and z are three elements of X s.t. fx $\sim$ fy, fy $\sim$ fz and fx>fz.

At least two states of S are not null states. Let $s_1$ and $s_2$ be two of the not null states and compare the decisions g, h, k:

g: $g(s_1) = x$,  $g(s_2) = y$,  $g(s) = x$ if $s \notin \{s_1, s_2\}$
h: $h(s_1) = z$,  $h(s_2) = x$,  $h(s) = x$ if $s \notin \{s_1, s_2\}$
k: $k(s_1) = y$,  $k(s_2) = z$,  $k(s) = x$ if $s \notin \{s_1, s_2\}$.

From P3: fx>fz implies that $fx_{|s1} > fz_{|s1}$

$fx \sim fy$ implies that $fx_{|s2} \sim fy_{|s2}$.

Since $fx_{|s1}=g(s_1)=x$ and $fy_{|s1}=h(s_1)=z$ : $g(s_1)>h(s_1)$.
Since $fy_{|s2}=g(s_2)=y$ and $fx_{|s2}=h(s_2)=x$ : $g(s_2) \sim h(s_2)$.

Thus, $[g \geq h] = S$ and $[h \geq g] = S - \{s_1\}$. Since $\{s_1\}>\emptyset$, applying auto-duality and the lifting axiom leads to obtain: g>h. Similarly, $[h \geq k]=S$ and $[k \geq h]=S-\{s_2\}$. Since $\{s_2\}>\emptyset$, we get h>k. Similarly, $[g \geq k] = S$ and $[k \geq g] = S$. Hence $g \sim k$.

Hence, assuming $fx \sim fy$, $fy \sim fz$ and fx>fz leads to get: g>h, h>k, g~k, which is in contradiction with the transitivity of >. Hence, ~ is transitive on X. □

**Property.**
$(A \cap (B \cup C)=\emptyset$ and A>B and A>C $\Rightarrow$ A>B$\cup$C); (6.15)
$(A \cap (B \cup C)=\emptyset$ and A>B and A$\geq$C $\Rightarrow$ A $\geq$ B$\cup$C). (6.16)

**Proof (6.15).** Let us suppose that A>B, A>C. From P5', there are three constant acts fx, fy and fz such as fx>fy>fz. Let us compare acts g,h,k:

|   | A | B$\cap$C | B$\cap\bar{C}$ | $\bar{B}\cap$C | $\bar{A}\cap\bar{B}\cap\bar{C}$ |
|---|---|---|---|---|---|
| g: | x | z | y | y | y |
| h: | y | y | x | y | y |
| k: | z | x | x | x | y |

From P3: $[g \geq h] = \bar{B}$ and $[h \geq g] = \bar{A}$. Since A>B, the lifting axiom and the autoduality property lead to: g>h (i)

From P3: $[h \geq k] = \bar{C}$ and $[k \geq h] = \bar{A}$. Since A>C, h>k (ii)

From P3: $[g \geq k] = S - (B \cup C)$ and $[k \geq g] = \bar{A}$.

Hence, (i), (ii) and the transitivity of > lead to: g>k, i.e., $S - (B \cup C) > \bar{A}$. Thus, applying the autoduality property: $A > B \cup C$.

Property 6.16 can be proved in the same way. □

Hence, the type of relation we obtain on S satisfies some of the characteristic properties of acceptance relations (cf. Section 5). But it is not limited to these acceptance relations (which are preorders on events).

Define now an inference relation as follows:

**Definition.** $A \mathrel{\vert\!\sim} B$ iff $A \cap B > A \cap \bar{B}$
where > is the relation defined by our axiomatic and projected on $2^S$ from $X^S$.

**Properties of $\mathrel{\vert\!\sim}$.** Restricted reflexivity, OR and AND follow from (P'1, P2, P3, P4, P5', L).

**Proofs.**
RR: Obvious.

AND: Consider the following acts

|   | ABC | AB$\bar{C}$ | A$\bar{B}$C | A$\bar{B}\bar{C}$ |
|---|---|---|---|---|
| f | x | y | z | z |
| g | y | z | x | y |
| h | z | x | y | x |

$[f>g] = A \cap B$ and $[f<g] = A \cap \bar{B}$: f>g
$[g>h] = A \cap C$, $[h>g] = A \cap \bar{C}$ : g>h

Since $[f>h] = A \cap B \cap C$ and $[h>f] = A \cap (\bar{C} \cup \bar{B})$ by transitivity of >: $A \cap B \cap C > A \cap (\bar{C} \cup \bar{B})$
OR is proved in the same way. □

So, some properties of preferential inference (system P) are recovered.

## 6.2 GETTING SYSTEM P

At this point of our axiomatisation, we cannot get all the properties of system P, but only some of them. In order to prove RW, CUT and CM, we need the property:

$B \cap C = \emptyset$ and $A \subseteq B$ and $A \geq C \Rightarrow B \geq C$

which cannot be derived from our set of axioms, except if there are no null events (pr. 6.10 and 6.11). Null events generally never appear in nonmonotonic reasoning nor belief revision theories. But supposing that there is no null event is not appropriate in the context of a Savage-like theory of decision. We prefer to add the unanimity axiom U proposed by Lehmann (1996):

**Axiom.** $f_{|A} \geq g_{|A}$ and $f_{|\bar{A}} \sim g_{|\bar{A}} \Rightarrow f \geq g$   U)

**Property.** If the relation on $X^S$ satisfies (P'1, P2, P3, P4, P'5, L, U), then
$B \cap C = \emptyset$ and $A \subseteq B$ and $A \geq C \Rightarrow B \geq C$   (6.17).

**Proof.** this property is verified if $B-A>\emptyset$ or A>C (Pr. 6.10 and 6.11). Suppose that A~C and B - A ~Ø.

Since $A \subseteq B$ and $B \cap C = \emptyset$: $B \subseteq \bar{C}$, i.e. $B - A = \bar{A} \cap B \subseteq \bar{A} \cap \bar{C}$

From P4, P5' and P2:
$A \sim C \Rightarrow \omega(A)^{x,y} \sim_{A \cup C} \omega(C)^{x,y}$
$B - A = \bar{A} \cap B \sim \emptyset \Rightarrow \omega(\bar{A} \cap B)^{x,y} \sim_{\bar{A} \cap \bar{C}} \omega(\emptyset)^{x,y}$

Let $f=\omega(B)^{x,y}$ and $g=\omega(C)^{x,y}$:
if $s \in A \cup C$, $f(s)=\omega(A)^{x,y}(s)$ and $g(s)=\omega(C)^{x,y}(s)$
if $s \in \bar{A} \cap \bar{C}$, $f(s)=\omega(\bar{A} \cap B)^{x,y}$ and $g(s)=\omega(\emptyset)^{x,y}(s)$.

Thus $f \sim_{A \cup C} g$ and $f \sim_{\bar{A} \cap \bar{C}} g$. Hence (B2): $f \sim g$, then, by P4, B~C □

**Property.** If the relation on $X^S$ satisfies (P'1, P2, P3, P4, P'5, L, U), then nonmonotonic inference built from the relation projected from $X^S$ to $2^S$ satisfies the properties of system P : OR, AND, RR, RW, CM, CUT.

**Sketches of proofs.**
RW can be easily proved from property 6.17
CM :

|   | A∩B∩C | A∩B∩C̄ | A∩B̄∩C | A∩B̄∩C̄ |
|---|---|---|---|---|
| f | x | y | y | z |
| g | y | z | x | y |
| h | z | x | y | x |

From A∩B > A∩B̄ , we get f>g. From A∩C > A∩C̄ we get g>h. By transitivity of >: f>h, i.e., A ∩ B ∩ C > A ∩ B ∩ C̄ ∪ A ∩ B̄ ∩ C. From property 6.17 (contraposition): A ∩ B ∩ C > A ∩ B ∩ C̄.
We can also obtain the CUT in the same way.    □

### 6.3 GETTING THE POSSIBILISTIC LIKEHOOD RELATION

So, starting from a general framework on acts, that respects Savage approach to a large extent, and adding a particular decision rule that tolerates a qualitative representation of uncertainty and preference we find a representation of uncertainty which is representable by a family of possibility measures, since the preferential entailment of system P can be always represented in terms of such a family (Dubois et al. 1995a). Notice in particular that the order of states obtained from our axiomatics is not necessarily a preorder, so that it cannot be represented by a single possibility ordering. The possibilistic likehood relation of Sections 3 and 4, based on a preorder on states and a necessity measure is recovered exactly if a counterpart of rational monotony is added to the above set of postulates. Indeed, adding the RM axiom forces the relation induced from the preference on acts to be a complete preordering, thus corresponding to a single possibilistic ordering as proved in (Benfehrat et. al 1992), adding rational monotony to system P. The possibilistic likehood relation in actually a refinement of this possibilic ordering for not Π-mutually exclusive events.

**Theorem.** The set of axioms (P'1, P2, P3, P4, P'5, L, RM, U ) is consistent.

**Proof.** Since the order on acts described in Section 3 and 4 satisfies all these axioms.

**Theorem:** If ($\geq$, $X^S$ ) satisfies (P'1, P2, P3, P4, P5'), the lifting axiom L, axioms U and RM, then there is a preorder $\geq_\pi$ on S and a preorder $\geq_P$ on X such as:
$$f \geq g \Leftrightarrow [f \geq_P g] \geq_N [g \geq_P f]$$
where N is the necessity ordering over events obtained from $\geq_\pi$ over states

**Sketch of proof.** A consequence of the lifting axiom (Pr. 6.14) we know that the projection of ($\geq$, $X^S$ ) on X is a preorder. Let us denote $\geq_P$ this preorder: [f $\geq_P$ g] = [f $\geq$ g]. From RM we also know that the projection ($\geq$, $X^S$ ) on S defines a preorder on states. Let us denote $\geq_\pi$ this preorder: s' $\geq_\pi$ s $\Leftrightarrow$ {s'} $\geq$ {s}.
From the autoduality property, its holds that B$\geq$A $\Leftrightarrow$ $\bar{A}\cap B \geq A \cap \bar{B}$. Hence, we have to compare disjoint events, which are mutually exlusive, i.e., ordered by the possibilistic ordering $\geq_\Pi$ corresponding to $\geq_\pi$. So : B$\geq$A $\Leftrightarrow$ $\bar{A}\cap B \geq_\Pi A\cap \bar{B}$ $\Leftrightarrow$ B$\geq_N$ A. With B= [f$\geq$g] and A = [g$\geq$f], the lifting axiom then leads to : [f$\geq$g]$\geq$[g$\geq$f] $\Leftrightarrow$ [f$\geq$g]) $\geq_N$ [g$\geq$f].    □

## 7 CONCLUSION

This result is rather negative for decision theory when only ordinal information about uncertainty and preference on consequences is available and no commensurability assumption is assumed between uncertainty and preference. Despite the presence of well-known Savage axioms, including the sure thing principle, the admissible uncertainty functions do not contain any kind of probability functions. This is clearly due to the lifting axiom.

The decision theory obtained captures either very risky attitudes or some that are not decisive at all. As preferential inference is very cautious, the relations on acts which do not correspond to a total ordering on states will not be very discriminating. On the contrary if the set of states is totally ordered in terms of plausibility, the decisions will be very risky because, as usual with rational inference the decision maker will always assume that the world is in the most normal state. Cautious decisions will never be preferred.

**Example.** Consider the omelette example of Savage (1972, pages 13 to 15). The problem is whether to add an egg to a 5-egg omelette: The set of 6 consequences is as in the following table:

| ACTS \ STATES | fresh egg | rotten egg |
|---|---|---|
| break the egg in the omelette BIO | a 6 egg omelette (6) | nothing to eat (1) |
| break it apart in a cup BAC | a 6 egg omelette, a cup to wash (5) | a 5 egg omelette, a cup to wash (3) |
| throw it away TA | a 5 egg omelette, one spoiled egg (2) | a 5 egg omelette (4) |

Integers between parentheses indicate the ordering of consequences, in decreasing order of preference. The reader can easily check that he agrees with this ordering. If fresh egg is more likely than rotten egg then A(BIO $\geq$ BAC) = A(BIO $\geq$ TA) = A(BAC $\geq$ TA) = {fresh} > A(BAC $\geq$ BIO) = A(TA $\geq$ BIO) = A(TA $\geq$ BAC) = {rotten}. So the decision making attitude induced by the approach is: break the egg in the omelette if you think the egg is fresh, throw it away if you think it is rotten, and do anything you like if you have no opinion (all acts equally preferred then). Clearly, this results in many starving days, and garbage cans with lots of spoiled fresh eggs.

Although the lifting axiom may look reasonable, and other axioms on acts innocuous, it is difficult to maintain that the decision guidelines offered by the theory are reasonable. In practice, it is advisable to act more cautiously, and to break the egg in a spare cup in case of serious doubt. In contrast the qualitative theory developed by Dubois and Prade (1995c) that justifies a pessimistic decision criterion generalizing Wald criterion looks more satisfactory. Applied to the egg example, it recommends act BAC in case of relative ignorance on the egg state; see (Dubois, Prade and Sabbadin, 1997). However that theory relies on a commensurability assumption between uncertainty and preference, which may be questioned by tenants of a purely symbolic approach.

This paper leads to an open question: Is there an alternative to the lifting axiom that would enable a reasonably cautious and albeit decisive ordering on acts to be computed on the basis of uncertainty relations on events and consequences? In other words, how to do away with commensurability assumptions and/or numerical approaches while capturing anthropomorphic decision attitudes?

One might think of reverting the lifting procedure by exchanging the role of states and consequences. It would lead to express the preference on acts in terms of a comparison of sets of consequences instead of sets of states as done here. This would make us capable of expressing attitudes of the decision maker in front of risk, something that never explicitly appears in the Savage's set of axioms. Indeed, a natural alternative to the lifting principle used in this paper would rather start by considering the set of consequences which are reached at least as certainly by decision $d_1$ as by decision $d_2$, namely the set of consequence $B(d_1 \geq d_2) = \left\{x, d_1^{-1}(x) \geq_U d_2^{-1}(x)\right\}$ where $d^{-1}(x)$ is the set of states in S from which d leads to consequence x. The problem is now to compare $B(d_1 \geq d_2)$ and $B(d_2 \geq d_1)$ in terms of preferences on X. Since $B(d_1 \geq d_2)$ is not usually a singleton on X, we need to add an additional hypothesis on DM's attitude in face of risk. Namely, if the DM is rather pessimistic (cautious), he/she may consider that $d_1 \geq d_2$ iff $\min_X B(d_1 \geq d_2) \geq_P \min_X B(d_2 \geq d_1)$, while if the DM is optimistic it may act on the basis of the $d_1 \geq d_2$ iff $\max_X B(d_1 \geq d_2) \geq_P \max_X B(d_2 \geq d_1)$. We may also think of comparing $\min_X B(d_1 \geq d_2)$ with $\max_X B(d_2 \geq d_1)$, or to refine the 'min' and 'max' over X by 'discrimin' and 'discrimax'. The investigation of this approach, which supposes that we know something about DM's attitude, is a topic for further research.